\documentclass[11pt]{article}
\usepackage{coling2020}
\usepackage{times}
\usepackage{url}
\usepackage{latexsym}

\usepackage{times}
\usepackage{url}
\usepackage{algorithm}
\usepackage{multicol}
\usepackage{algorithmicx}
\usepackage{mathrsfs}
\usepackage{latexsym}
\usepackage{graphicx}
\usepackage{float}
\usepackage{amsmath, bm}
\usepackage{array}
\usepackage{bbding}

\colingfinalcopy 

\title{KnowDis: Knowledge Enhanced Data Augmentation for Event Causality Detection via Distant Supervision}

\author{Xinyu Zuo$^{1,2}$, Yubo Chen$^{1,2}$, Kang Liu$^{1,2}$, Jun Zhao$^{1,2}$ \\
	$^1$National Laboratory of Pattern Recognition, Institute of Automation, \\
	Chinese Academy of Sciences, Beijing, 100190, China \\
	$^2$School of Artificial Intelligence, \\
	University of Chinese Academy of Sciences, Beijing, 100049, China \\
	{\tt \{xinyu.zuo,yubo.chen,kliu,jzhao\}@nlpr.ia.ac.cn}}

\date{}

\begin{document}
	\maketitle
	\begin{abstract}
		Modern models of event causality detection (ECD) are mainly based on supervised learning from small hand-labeled corpora. However, hand-labeled training data is expensive to produce, low coverage of causal expressions and limited in size, which makes supervised methods hard to detect causal relations between events. To solve this data lacking problem, we investigate a data augmentation framework for ECD, dubbed as \textbf{Know}ledge Enhanced \textbf{Dis}tant Data Augmentation (\textbf{KnowDis}). 
		Experimental results on two benchmark datasets \emph{EventStoryLine corpus} and \emph{Causal-TimeBank} show that 1) KnowDis can augment available training data assisted with the lexical and causal commonsense knowledge for ECD via distant supervision, and 2) our method outperforms previous methods by a large margin assisted with automatically labeled training data.
	\end{abstract}
	
	\section{Introduction}
	\label{intro}
	Event causality detection (ECD) aims to identify causal relations between events from texts, which may provide crucial clues for many NLP tasks, such as information extraction, logical reasoning, question answering, and others \cite{girju2003automatic,oh2013question,oh2017multi}. 
	For example, the causal relation that Kimani Gray was \emph{killed} because of a police \emph{attack} is needed to be detected in the following sentence: \emph{"Kimani Gray, a young man who likes football, was \textbf{killed} in a police \textbf{attack} shortly after a tight match."}
	
	This task is usually modeled as a classification problem, i.e. determining whether there is a causal relation between two events in a sentence. To this end, most existing methods adopt a supervised learning paradigm \cite{mirza2016catena,riaz2014recognizing,hashimoto2014toward,hu2017inferring,gao2019modeling,zuo2020causal}. Although these methods have achieved good performance, they usually need large-scale annotated training data. However, existing event causality detection datasets are relatively small. For example, the EventStoryLine Corpus \cite{caselli2017event} only contains 258 documents, 4316 sentences, and 1770 causal event pairs. These small datasets are in low coverage of causal expressions and obstacle NLP applications deployed on large-scale data.  Recent improvements of distant supervision have been proven to be effective to label training data for some tasks, such as relation extraction \cite{Mintz2009DistantSF}, event detection \cite{chen-etal-2017-automatically}, and so on. Therefore, we investigate a distant data augmentation framework for solving the data lacking problem on the ECD task, dubbed as \textbf{Know}ledge Enhanced \textbf{Dis}tant Data Augmentation (\textbf{KnowDis}), to automatically label available data.
	
	We argue that a sentence contains an event pair with a high probability of causality and expresses its causal semantic can be labeled as training data for the ECD task. To automatically label a large number of training data, we need to solve the following three challenges. (1) How to collect a large number of event pairs with a high probability of causality and employ them to label training data. (2) How to handle noisy distantly labeled sentences that do not have well-formed textual expressions to express causal semantics. (3) How to make better use of distantly labeled sentences for training. To this end, we firstly design a \textbf{Lexi}con Enhanced \textbf{Anno}tator (\textbf{LexiAnno}) to extract a large number of event pairs with a high probability of causality based on lexical knowledge and employ them to automatically label sentences via distant supervision. Secondly, we propose a \textbf{Common}sense \textbf{Filter} (\textbf{CommonFilter}) to refine distantly labeled sentences assisted with causal commonsense knowledge which makes them more well-formed to express the causal semantics. Thirdly, we employ \textbf{Relabeling} and \textbf{Annealing} strategies to make better use of distantly labeled sentences for training. Finally, we evaluate KnowDis on two datasets and achieve the best performance training with distantly labeled sentences on ECD. The following sections describe the architecture (Section \ref{KnowDis}) of KnowDis and the experimental results (Section \ref{Experiment}) on the ECD task.
	
	\section{KnowDis}
	\label{KnowDis}
	\begin{figure*}[t] \footnotesize
		\centering
		\includegraphics*[clip=true,width=0.70\textwidth,height=0.16\textheight]{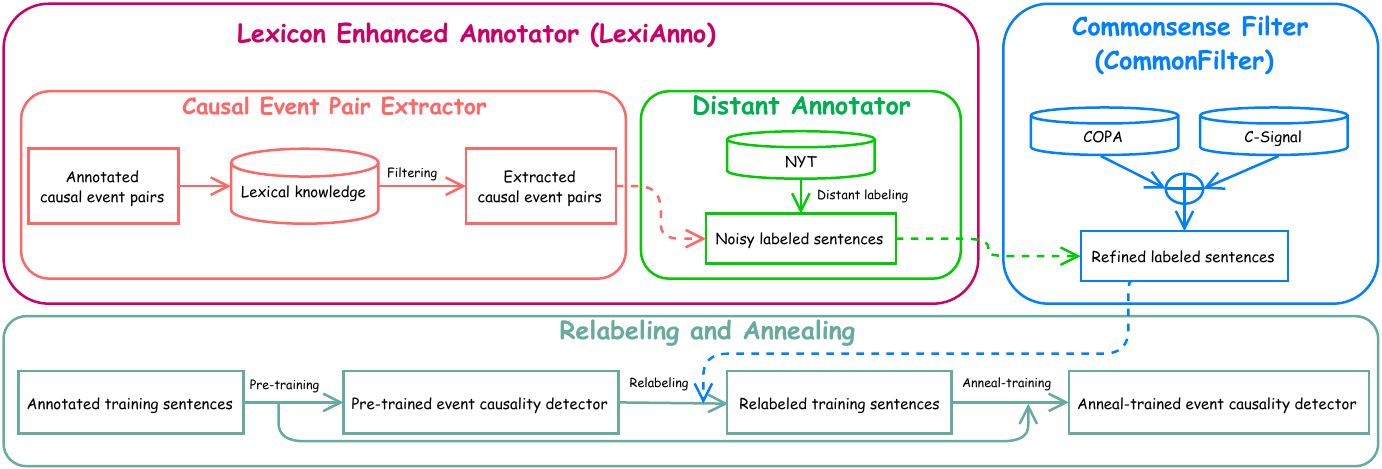}
		\caption{Overview of Knowledge Enhanced Distant Data Augmentation framework (KnowDis).} \label{fig2}
	\end{figure*}
	
	As shown in Figure \ref{fig2}, we illustrate the three main components of our proposed KnowDis in this section. 
	
	\subsection{Lexicon Enhanced Annotator (LexiAnno)}
	\label{LexiAnno}
	LexiAnno aims to extract a large number of event pairs with a high probability from external lexicons based on the annotated causal event pairs via a \emph{Causal Event Pair Extractor}, and employ them to collect preliminary noisy labeled sentences from external documents via a \emph{Distant Annotator}.
	
	\begin{table*}[h] \footnotesize
		\centering
		\scalebox{.85}{
			\begin{tabular}{|m{1.5cm}|m{8cm}|m{4cm}|m{0.5cm}|}
				\hline
				\multicolumn{1}{|c|}{\textbf{Knowledge}}  & \multicolumn{1}{c|}{\textbf{How to extract}}    & \multicolumn{1}{c|}{\textbf{Why causality}}     & \multicolumn{1}{|c|}{\textbf{Abbr.}}  \\ \hline
				\multicolumn{1}{|c|}{\textbf{WordNet}}  & 1) Extracting the synonyms and hypernyms from WordNet of head word of each event in $e_{ij}$. 2) Assembling the items from the two groups of two events to generate causal event pair set.                                                                                                                                                                                                                                                                                                                                                                                                                                                                                                                 & Items in each group are the synonyms and hypernyms of the original causal event pairs. & \multicolumn{1}{|c|}{$E^{wn}$}       \\ \hline
				\multicolumn{1}{|c|}{\textbf{VerbNet}}                                                                    & 1) Extracting the words from VerbNet under the same class as head word of each event in $e_{ij}$. 2) Assembling the items from the two groups of two events to generate causal event pair set.                                                                                                                                                                                                                                                                                                                                                                                                                                                                                                        & Items in each group are in the same class of the original causal event pairs.  & \multicolumn{1}{|c|}{$E^{vn}$}       \\ \hline
		\end{tabular}}
		\caption{Extracting causal event pairs from lexical knowledge bases.}
		\label{tab1}
	\end{table*}
	
	\textbf{Causal Event Pair Extractor.} We expand each event pair $e_{ij}$ in annotated causal event pair set $E^{g}$ via external dictionaries.\footnote{\emph{WordNet}: https://wordnet.princeton.edu/ and \emph{VerbNet}: http://verbs.colorado.edu/~mpalmer/projects/verbnet.html}. Table \ref{tab1} illustrates the details of how to extract $E^{wn}$ and $E^{vn}$ from WordNet \cite{miller1995wordnet} and VerbNet \cite{schuler2005verbnet}. Eventually, we construct a filter via transE \cite{bordes2013translating} based on maximum interval method: $L=\sum_{(e_i,e_j) \in S} \sum_{(e'_i,e'_j) \in S'} [\lambda + d(\bm{e'_i},\bm{e'_j})-d(\bm{e_i},\bm{e_j})]_{+}$ to sort extracted event pairs in ascending order of their distance and pick out the ones at the top of them with a high probability of causality, where $S$ and $S'$ are the causal and non-causal event pair set respectively. 
	
	\textbf{Distant Annotator.} We keep the top 10\% sorted extracted event pairs to obtain $E^{f}$ with a high probability of causality. Then we automatically label the 5\% randomly selected sentences from NYT corpus\footnote{NYT corpus is very large, so we randomly select 5\% sentences from NYT corpus as the sentences to be labeled.}	which contain any event pair $e_{ij}$ in $E^{g}$ and $E^{f}$ as the noisy distantly labeled training data $\bm{D_{n}}$. 
	
	\subsection{Commonsense Filter (CommonFilter)}
	\label{CommonFilter}
	\begin{table*}[h] \footnotesize
		\centering
		\scalebox{.85}{
			\begin{tabular}{|m{1.5cm}|m{6.5cm}|m{2.8cm}|m{2.8cm}|}
				\hline
				\multicolumn{1}{|c|}{\textbf{Source}}  & \multicolumn{1}{c|}{\textbf{Data Form}}    & \multicolumn{1}{c|}{\textbf{Cause-related Text}}     & \multicolumn{1}{c|}{\textbf{Effect-related Text}}  \\ \hline
				\multicolumn{1}{|c|}{\textbf{COPA}}                                                                    & \emph{Premise}: The woman hired a lawyer. 
				
				\emph{Alternative1}: She decided to sue her employer. (\Checkmark) 
				
				\emph{Alternative2}: She decided to run for office. (\XSolidBrush) & She decided to sue her employer &  The woman hired a lawyer      \\ \hline
				\multicolumn{1}{|c|}{\textbf{Annotated Data}}                                                                    & Kimani Gary, a young man who likes football, was killed in a police attack shortly after a tight match.                                                                                                                                                                                                                                                                                                                                                                                                                                                                                                        &   a police attack shortly after a tight match    &  Kimani Gary, a young man who likes football, was killed      \\ \hline
		\end{tabular}}
		\caption{Cause-related and effect-related text from COPA and annotated data.}
		\label{tab2}
	\end{table*}
	
	CommonFilter aims to refine $\bm{D_{n}}$ assisted with causal commonsense knowledge to pick out labeled sentences which express causal semantics between events. Inspired by Luo et al. \shortcite{Luo2016CommonsenseCR}, we introduce Pointwise Mutual Information (PMI) statistics \cite{Church1989WordAN} to indicate the causal semantics assisted with a \emph{if-then} reasoning data Choice of Plausible Alternatives (COPA) \cite{Gordon2011SemEval2012T7} and causal connectives. As shown in Table \ref{tab2}, we employ causality co-occurrences ($f$) of each word pair between cause-related ($T_c$) and effect-related ($T_e$) text and incorporate \emph{necessity causality} ($CS_{nec}$) with \emph{sufficiency causality} ($CS_{suf}$) to model causal relation. Specifically, we calculate the $CS_{nec}$ and $CS_{suf}$ of each word pair ($i_c$, $j_e$) in ($T_c$, $T_e$) from COPA and annotated data, and causality score $CS_s$ of two text spans ($SP_1$, $SP_2$) of each sentence $s$ in $\bm{D_{n}}$ divided with connectives between two events:
	
	\begin{equation} \footnotesize
	CS_{nec}(i_c,j_e)=\frac{p(i_c|j_e)}{p^{\alpha}(i_c)}=\frac{p(i_c,j_e)}{p^{\alpha}(i_c)p(j_e)}, CS_{suf}(i_c,j_e)=\frac{p(j_e|i_c)}{p^{\alpha}(j_e)}=\frac{p(i_c,j_e)}{p(i_c)p^{\alpha}(j_e)}
	\vspace{-10pt}
	\end{equation}
	
	\begin{equation} \footnotesize
	p(i_c)=\frac{\sum_{w \in W}f(i_c,w_e)}{M}, 
	p(j_e)=\frac{\sum_{w \in W}f(w_c,j_e)}{M},
	p(i_c,j_e)=\frac{f(i_c,j_e)}{N},
	M=\sum_{u \in W}\sum_{v \in W}f(u_c, v_e),
	\vspace{-8pt}
	\end{equation}
	
	\begin{equation} \footnotesize
	CS(i_c,j_e) = CS_{nec}(i_c,j_e)^\lambda CS_{suf}(i_c,j_e)^{1-\lambda}, CS_s(SP_1, SP_2)=\frac{1}{|SP_1|+|SP_2|}\sum_{i \in SP_1}\sum_{j \in SP_2}CS(i,j)
	\end{equation}
	where, $N$ is the size of all ($T_c$, $T_e$) pairs, $W$ is all calculated words and $\alpha$ is a penalty value to penalize high-frequency words. Next, we sort and divide sentences in $\bm{D_{n}}$ into two parts based on $CS_s$, $\bm{D_n^c}$ in which the two events are connected by a causal connective from $\bm{C_{signal}}$ extracted from FrameNet \cite{Baker1998TheBF} and PDTB2 \cite{pdtb2008pdtb}, and the $\bm{D_n^{nc}}$ in which are not. Finally, we keep the top 50\% data in $\bm{D_n^c}$ and 10\% data in $\bm{D_n^{nc}}$ as refined distantly labeled training data $\bm{D_r}$. 
	
	\subsection{Relabeling and Annealing}
	\label{Training}
	\textbf{Event Causality Detector.} We formulate event causality detection as a sentence-level binary classification problem. Specifically, we design a binary classifier based on BERT \cite{devlin-etal-2019-bert} to construct the \emph{Event Causality Detector}. The input of the detector is the event pair $e_{ij}$ and its corresponding sentence. We convert the sentence into BERT's input form, i.e. the sum of WordPiece embedding \cite{wu2016google}, position embedding, and segment embedding. We get the event representation $\bm{e}_i$ and $\bm{e}_j$ encoded by BERT. Then, we take the stitching of manual designed feature vector (same lexical, causal potential, and syntactic features representation as Gao et al. \cite{gao2019modeling}) $\bm{f}$, $\bm{e}_i$ and $\bm{e}_j$ as the input of top MLP classifier. Finally, the output is a binary vector to indicate the causality of the input event pair $e_{ij}$.
	
	We employ relabeling and annealing strategies to make better use of distantly labeled data for training. (1) \emph{Relabeling}: We pre-train a detector on annotated data and employ it to relabel the refined distantly labeled training data $\bm{D_r}$ via self-training  \cite{Asai2020LogicGuidedDA}. Then, we collect the sentences that are relabeled as causal sentences to obtain the distantly relabeled training data $\bm{D_{rr}}$ which are more casual and informative for the training of ECD task. (2) \emph{Annealing}: Distantly labeled training data may not be appropriate at the beginning of training for building an effective detector due to noises. Therefore, we employ the annealing training strategy \cite{Kirkpatrick1983OptimizationBS} to maximize the effectiveness of distantly labeled training data. In the beginning, we only employ annotated data for training, and with the increase of epochs, we added $\bm{D_{rr}}$ for training incrementally in a proportion of $\beta$.
	
	\section{Experiments}
	\label{Experiment}	
	
	\textbf{Datasets.}: (1) \textbf{ESC}: We use the same way to partition dataset as the SOTA method on ESC \cite{gao2019modeling}. Same as it, we use the last two topics as a development set. (2) \textbf{Causal-TB}: This dataset only contains 318 causal links which can further prove effectiveness of the proposed framework for solving the problem of data lacking. We use the same development set as ESC because of the SOTA method on this dataset \cite{Mirza2014AnAO} does not partition the development set. Specifically, we conduct 5-fold cross-validation on the two datasets\footnote{For each fold, we add extra distantly labeled data based on the annotated event pairs corresponding to this fold for training.}. We tune the augmented proportion, $\alpha$, and $\beta$ on the development set. All the results are the average of three independent experiments. 
	
	\textbf{Parameters Setting.} We apply the \emph{base-uncase-bert} as the pre-trained BERT model. We set the learning rate of detector as 1e-5. Specifically, the dimension of the causal semantic space is 100. We set the $\alpha$ and $\beta$ as 0.5 and 0.1 respectively based on the development set. We apply the early stop strategy and the SGD gradient strategy to optimize all models. We adopt \emph{Precision} (P), \emph{Recall} (R), \emph{F1 vaule} (F1) as the evaluation metrics.
	
	\textbf{Compared Methods.} We evaluate the performance of ECD on the same EventStoryLine corpus v0.9 (ESC) \cite{caselli2017event} and Causal-TimeBank (Causal-TB) \cite{Mirza2014AnAO} dataset as SOTA methods. We select some typical methods and SOTA methods on ESC and Causal-TB respectively to make comparisons: (1) \textbf{Cheng et al. \shortcite{cheng2017classifying}} and \textbf{Choubey et al. \shortcite{choubey2017sequential}}: two dependency path based sequential neural models which have shown effectiveness on ESC. (2) \textbf{Gao et al. \shortcite{gao2019modeling}}: the SOTA method which models the document-level structures for event causality detection on ESC. (3) \textbf{Mirza et al. \shortcite{Mirza2014AnAO}}: a strong supervised classifier with gold causal signals on Causal-TB. (4) \textbf{BERT}: our proposed detector (\ref{Training}), a strong baseline for comparison which performs significantly well in classification tasks. The other compared methods are not open source entirely, so we construct a BERT-based model as a strong baseline to verify our data augmentation framework. (5) \textbf{EDA}: training detector (\ref{Training}) with extra data augmented by EDA which is a easy data augmentation framework \cite{wei-zou-2019-eda}. We also employ relabeling and annealing strategy when training with EDA. Finally, we automatically label 10132 sentences via KnowDis. We sample 100 sentences for manual evaluation, 82\% of which clearly express the causal semantics (3 assessors, Cohen's kappa = 0.88). 
	
	\subsection{Comparisons with SOTA Methods on Event Causality Detection}
	\begin{minipage}{\textwidth}
		\centering
		\begin{minipage}[t]{0.45\textwidth}
			\centering
			\makeatletter\def\@captype{table}\makeatother
			\scalebox{.70}{
				\begin{tabular}{|c|c|c|c|c|}
					\hline
					\multicolumn{5}{|c|}{\textbf{ESC}} \\ \hline
					\textbf{Methods} & \textbf{P} & \textbf{R} & \textbf{F1} & \bm{$\nabla$}  \\ \hline
					\textbf{Cheng et al.\shortcite{cheng2017classifying}}     & 34.0    & 41.5    & 37.4   & -  \\ \hline
					\textbf{Choubey et al.\shortcite{choubey2017sequential}}     & 32.7    & 44.9    & 37.8   & -  \\ \hline
					\textbf{Gao et al.\shortcite{gao2019modeling}}     & 37.4   & 55.8    & 44.7    & -     \\ \hline
					\textbf{BERT}   & 36.6   & 59.7   & 45.3   & +0.6            \\ \hline 
					\multicolumn{5}{|c|}{}  \\[-12pt]\hline
					\textbf{EDA\cite{wei-zou-2019-eda}} &  38.8 &  62.9  &  48.0 &  +3.3  \\ \hline
					\textbf{KnowDis (our method)}  & 39.7    & 66.5   & \textbf{49.7}   & +5.0    \\ \hline
			\end{tabular}}
			\caption{Performance of compared methods for ECD on ESC. \bm{$\nabla$} means the points higher than the SOTA method on F1 value.}
			\label{tab3}
		\end{minipage}
		\begin{minipage}[t]{0.45\textwidth}
			\centering
			\makeatletter\def\@captype{table}\makeatother
			\scalebox{.70}{
				\begin{tabular}{|c|c|c|c|c|}
					\multicolumn{5}{c}{}\\
					\multicolumn{5}{c}{}\\
					\hline
					\multicolumn{5}{|c|}{\textbf{Causal-TB}}\\ \hline
					\textbf{Methods} & \textbf{P} & \textbf{R} & \textbf{F1} & \bm{$\nabla$} \\ \hline
					\textbf{Mirza et al. \shortcite{Mirza2014AnAO}}  &  74.6  &  35.2  & 47.8 & - \\ \hline
					\textbf{BERT} & 39.0  & 60.5  &  47.4 &  -0.4            \\ \hline
					\multicolumn{5}{|c|}{}  \\[-12pt]\hline
					\textbf{EDA\cite{wei-zou-2019-eda}} & 40.2  &  61.2  &  48.5  &  +0.7 \\ \hline
					\textbf{KnowDis (our method)}  &   42.3    &  60.5   &  \textbf{49.8}   &   +2.0     \\ \hline
			\end{tabular}}
			\caption{Performance of compared methods for ECD on Causal-TB. \bm{$\nabla$} means the points higher than the SOTA method on F1 value.}
			\label{tab4}
		\end{minipage}
		\vspace{+10pt}
	\end{minipage}
	
	Table \ref{tab3} and \ref{tab4} shows the results of our model compared with SOTA methods. From the results, we could have the main following observations. (1) \textbf{\emph{Effectiveness of our method}}: Our method (\textbf{KnowDis}) significantly improves the performance of ECD by 5.0 and 2.0 points on F1 value on two datasets respectively. It illustrates that the augmented training data labeled via distant supervision, and refined via causal commonsense knowledge can provide more effective assistance for ECD task. (2) \textbf{\emph{Necessity of causal-related knowledge}}: We can observe that training with extra data augmented with \textbf{EDA} and \textbf{KnowDis} can both improve the performance of ECD task which shows that more training data can introduce more causal knowledge to alleviate data scarcity. However, the sentences produced via \textbf{EDA} are not refined by causal-related knowledge such as causal lexical and causal commonsense knowledge. Compared to it, the further significant improvement of \textbf{KnowDis} proves the necessity of the causal-related knowledge, and also illustrates that our model can produce more suitable training data for the ECD task.
	
	\vspace{-4pt}
	\subsection{Effectiveness of Main Components on Event Causality Detection}
	\begin{minipage}{\textwidth}
		\centering
		\begin{minipage}[t]{0.45\textwidth}
			\centering
			\makeatletter\def\@captype{table}\makeatother
			\scalebox{.73}{
				\begin{tabular}{|c|cccc|}
					\multicolumn{5}{c}{}\\
					\hline
					\multicolumn{1}{|c|}{\textbf{Method}} & \multicolumn{1}{c}{\textbf{P}} & \multicolumn{1}{c}{\textbf{R}} & \multicolumn{1}{c}{\textbf{F}}    & \bm{$\nabla$}        \\ \hline
					\multicolumn{1}{|l|}{\textbf{BERT} (Baseline)}   & 36.6  & 59.7   & 45.3  & -      \\
					\multicolumn{1}{|l|}{\textbf{+Design features}} & 36.8 & 63.0  & 46.5 & +1.2  \\
					\multicolumn{1}{|l|}{\textbf{+Annotated causal ep.}$\bm{\dag}$ } & 37.5  & 67.0  & 48.1 & +2.8 \\
					\multicolumn{1}{|l|}{\textbf{+Extracted causal ep.}$\bm{\dag}$ } &  39.7    &  66.5   & \textbf{49.7}   & +4.4   \\ \hline
			\end{tabular}}
			\caption{Effectiveness of LexiAnno (\ref{LexiAnno}) and Detector (\ref{Training}) on ESC.}
			\label{tab5}
		\end{minipage}
		\begin{minipage}[t]{0.45\textwidth}
			\centering
			\makeatletter\def\@captype{table}\makeatother
			\scalebox{.73}{
				\begin{tabular}{|c|cccc|}
					\hline
					\multicolumn{1}{|c|}{\textbf{Method}} & \multicolumn{1}{c}{\textbf{P}} & \multicolumn{1}{c}{\textbf{R}} & \multicolumn{1}{c}{\textbf{F}}    & \bm{$\nabla$}   \\ \hline
					\multicolumn{1}{|l|}{\textbf{KnowDis} (Our model)}   & 39.7   & 66.5   & \textbf{49.7}   & -           \\
					\multicolumn{1}{|l|}{\textbf{-Causal connective}}  & 37.3  & 67.9  & 48.1   & -1.6  \\
					\multicolumn{1}{|l|}{\textbf{-Causality co-occurrence}} & 36.7       & 67.2       & 47.5 & -2.2 \\
					\multicolumn{1}{|l|}{\textbf{-Relabeling}} & 38.1       & 67.0       & 48.6 & -1.1         \\ 
					\multicolumn{1}{|l|}{\textbf{-Annealing}} & 37.6       & 67.8       & 48.4 & -1.3 \\ \hline
			\end{tabular}}
			\caption{Effectiveness of CommonFilter (\ref{CommonFilter}), Relabeling and Annealing (\ref{Training}) on ESC.}
			\label{tab6}
		\end{minipage}
		\vspace{+10pt}
	\end{minipage}
	
	Table \ref{tab5} and \ref{tab6} tries to show the effectiveness of the key parts of our method on event causality detection (ECD). $\bm{\dag}$ denotes the same filtering and training processes except that the causal event pairs employed for distant labeling are different. From the results, we could have the following observations. (1) \textbf{\emph{Effectiveness of distant labeling}}: The results of \textbf{Annotated causal ep.}$\bm{\dag}$ and \textbf{+Extracted causal ep.}$\bm{\dag}$ illustrates that the distantly labeled augmented training data can effectively alleviate the problem of data scarcity on ECD task. (\ref{LexiAnno}) (2) \textbf{\emph{Effectiveness of LexiAnno}}: Compared to sentences labeled only based on annotated causal event pairs (\textbf{+Annotated causal ep.}$\bm{\dag}$), training with sentences labeled based on extracted causal event pairs from knowledge bases (\textbf{+Extracted causal ep.}$\bm{\dag}$) can bring effective and diverse knowledge for understanding event-causal semantics. (\ref{LexiAnno}) (3) \textbf{\emph{Effectiveness of CommonFilter}}: The results of \textbf{-Causal connective} and \textbf{-Causality co-occurrence} show that our proposed commonsense filter (\ref{CommonFilter}) which introduces causal commonsense knowledge to refine distantly labeled training data can effectively enhance the causal-related semantics of augmented data. Specifically, the causal commonsense knowledge is more useful than causal connective knowledge because the former is more extensive than the latter in the expression of cause and effect. (4) \textbf{\emph{Effectiveness of Relabeling}}: The results of \textbf{-Relabeling} show that relabeling (\ref{Training}) can reduce the noisy of distantly labeled data. (5) \textbf{\emph{Effectiveness of Annealing}}: The results of \textbf{-Annealing} show that the annealing (\ref{Training}) can make better use of noisy distantly labeled data. Relabeling and annealing can both be applied to other distant supervision tasks. 
	
	\subsubsection{Effectiveness of Different Proportion of Distantly Labeled Data}
	\begin{figure*}[h]
		\centering
		\begin{minipage}[t]{0.48\textwidth}
			\centering
			\includegraphics[width=2in]{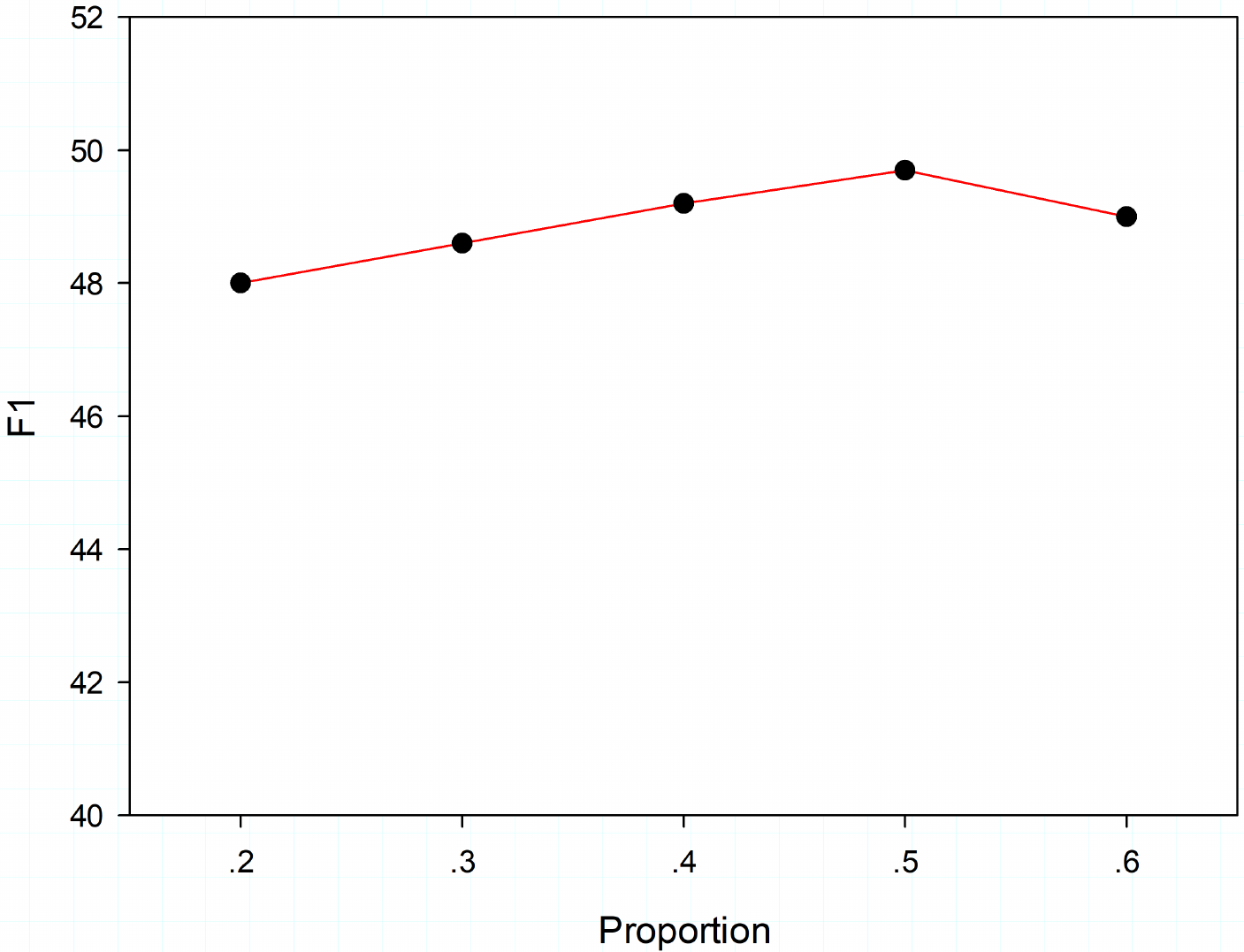}
			\caption{Effectiveness of different proportion of retained distantly labeled data in $\bm{D_n^c}$ on ESC.} \label{fig3}
		\end{minipage}
		\begin{minipage}[t]{0.48\textwidth}
			\centering
			\includegraphics[width=2in]{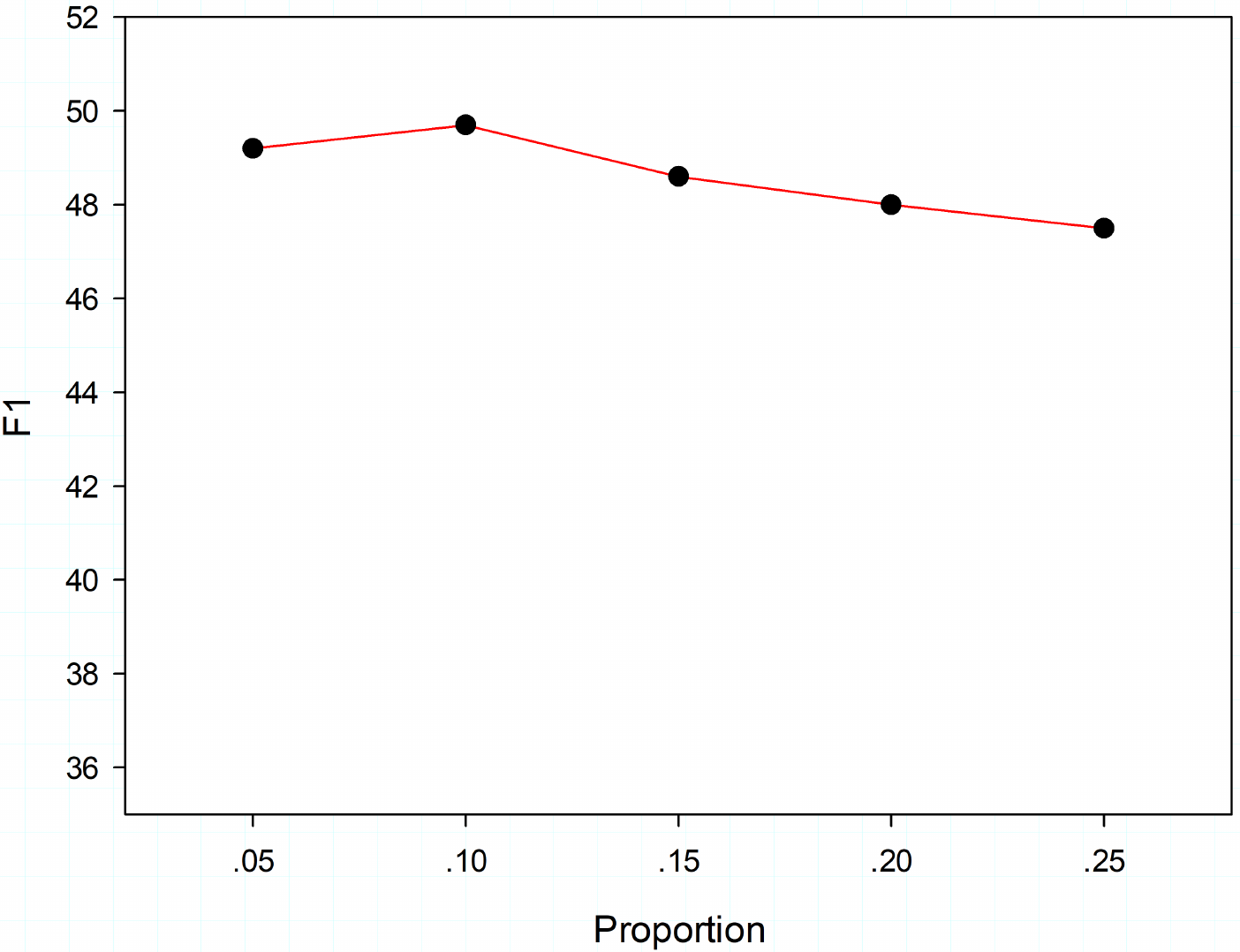}
			\caption{Effectiveness of different proportion of retained distantly labeled data in $\bm{D_n^{nc}}$ on ESC.} \label{fig4}
		\end{minipage}
	\end{figure*}
	
	Figure \ref{fig3} and \ref{fig4} tries to show the effectiveness of different proportion of retained distantly labeled data in $\bm{D_n^c}$ and $\bm{D_n^{nc}}$. From the results, we could have the main following observations. (1) 
	The more data retained of the distant label data in $\bm{D_n^c}$, the more effective knowledge can be brought for training. However, when the retained data exceeds 50\%, the noise caused by $\bm{D_n^c}$ is greater than the impact of effective knowledge. (2) Introducing appropriate distant label data in $\bm{D_n^{nc}}$ can bring additional effective knowledge but it contains more harmful noise than the data in $\bm{D_n^c}$.
	
	\section{Conclusion}
	In this paper, we try to employ distant supervision to alleviate the data lacking problem on causal-related task. We propose a knowledge enhanced distant data augmentation framework (KnowDis) for event causality detection. 
	Our method achieves the SOTA performance on EventStoryLine corpus and Causal-TimeBank dataset assisted with knowledge enhanced distantly labeled training data. In the future, we will introduce more causal-related resources and apply KnowDis for other relational tasks.
	
	\section*{Acknowledgements}
	This work is supported by the National Key R\&D Program of China (2020AAA0106400), the National Natural Science Foundation of China (No.61533018, No. 61922085, No.61806201) and the Key Research Program of the Chinese Academy of Sciences (Grant NO. ZDBS-SSW-JSC006). This work is also supported by a grant from Ant Group and Beijing Academy of Artificial Intelligence (BAAI).
	
	\bibliographystyle{coling}
	\bibliography{coling2020}

\begin{thebibliography}{}

\bibitem[\protect\citename{Asai and Hajishirzi}2020]{Asai2020LogicGuidedDA}
Akari Asai and Hannaneh Hajishirzi.
\newblock 2020.
\newblock Logic-guided data augmentation and regularization for consistent
  question answering.
\newblock {\em ArXiv}, abs/2004.10157.

\bibitem[\protect\citename{Baker \bgroup et al.\egroup }1998]{Baker1998TheBF}
Collin~F. Baker, Charles~J. Fillmore, and John~B. Lowe.
\newblock 1998.
\newblock The berkeley framenet project.
\newblock In {\em COLING-ACL}.

\bibitem[\protect\citename{Bordes \bgroup et al.\egroup
  }2013]{bordes2013translating}
Antoine Bordes, Nicolas Usunier, Alberto Garcia-Duran, Jason Weston, and Oksana
  Yakhnenko.
\newblock 2013.
\newblock Translating embeddings for modeling multi-relational data.
\newblock In {\em Advances in neural information processing systems}, pages
  2787--2795.

\bibitem[\protect\citename{Caselli and Vossen}2017]{caselli2017event}
Tommaso Caselli and Piek Vossen.
\newblock 2017.
\newblock The event storyline corpus: A new benchmark for causal and temporal
  relation extraction.
\newblock In {\em Proceedings of the Events and Stories in the News Workshop},
  pages 77--86.

\bibitem[\protect\citename{Chen \bgroup et al.\egroup
  }2017]{chen-etal-2017-automatically}
Yubo Chen, Shulin Liu, Xiang Zhang, Kang Liu, and Jun Zhao.
\newblock 2017.
\newblock Automatically labeled data generation for large scale event
  extraction.
\newblock In {\em Proceedings of the 55th Annual Meeting of the Association for
  Computational Linguistics (Volume 1: Long Papers)}, pages 409--419,
  Vancouver, Canada, July. Association for Computational Linguistics.

\bibitem[\protect\citename{Cheng and Miyao}2017]{cheng2017classifying}
Fei Cheng and Yusuke Miyao.
\newblock 2017.
\newblock Classifying temporal relations by bidirectional lstm over dependency
  paths.
\newblock In {\em Proceedings of the 55th Annual Meeting of the Association for
  Computational Linguistics (Volume 2: Short Papers)}, pages 1--6.

\bibitem[\protect\citename{Choubey and Huang}2017]{choubey2017sequential}
Prafulla~Kumar Choubey and Ruihong Huang.
\newblock 2017.
\newblock A sequential model for classifying temporal relations between
  intra-sentence events.
\newblock {\em arXiv preprint arXiv:1707.07343}.

\bibitem[\protect\citename{Church and Hanks}1989]{Church1989WordAN}
Kenneth~Ward Church and Patrick Hanks.
\newblock 1989.
\newblock Word association norms, mutual information and lexicography.
\newblock In {\em ACL}.

\bibitem[\protect\citename{Devlin \bgroup et al.\egroup
  }2019]{devlin-etal-2019-bert}
Jacob Devlin, Ming-Wei Chang, Kenton Lee, and Kristina Toutanova.
\newblock 2019.
\newblock {BERT}: Pre-training of deep bidirectional transformers for language
  understanding.
\newblock In {\em Proceedings of the 2019 Conference of the North {A}merican
  Chapter of the Association for Computational Linguistics: Human Language
  Technologies, Volume 1 (Long and Short Papers)}, pages 4171--4186,
  Minneapolis, Minnesota, June. Association for Computational Linguistics.

\bibitem[\protect\citename{Gao \bgroup et al.\egroup }2019]{gao2019modeling}
Lei Gao, Prafulla~Kumar Choubey, and Ruihong Huang.
\newblock 2019.
\newblock Modeling document-level causal structures for event causal relation
  identification.
\newblock In {\em Proceedings of the 2019 Conference of the North American
  Chapter of the Association for Computational Linguistics: Human Language
  Technologies, Volume 1 (Long and Short Papers)}, pages 1808--1817.

\bibitem[\protect\citename{Girju}2003]{girju2003automatic}
Roxana Girju.
\newblock 2003.
\newblock Automatic detection of causal relations for question answering.
\newblock In {\em Proceedings of the ACL 2003 workshop on Multilingual
  summarization and question answering-Volume 12}, pages 76--83. Association
  for Computational Linguistics.

\bibitem[\protect\citename{Gordon \bgroup et al.\egroup
  }2011]{Gordon2011SemEval2012T7}
Andrew~S. Gordon, Zornitsa Kozareva, and Melissa Roemmele.
\newblock 2011.
\newblock Semeval-2012 task 7: Choice of plausible alternatives: An evaluation
  of commonsense causal reasoning.
\newblock In {\em SemEval@NAACL-HLT}.

\bibitem[\protect\citename{Group and others}2008]{pdtb2008pdtb}
PDTB~Research Group et~al.
\newblock 2008.
\newblock The pdtb 2.0.
\newblock {\em Annotation Manual. Technical Report IRCS-08-01, Institute for
  Research in Cognitive Science, University of Pennsylvania}.

\bibitem[\protect\citename{Hashimoto \bgroup et al.\egroup
  }2014]{hashimoto2014toward}
Chikara Hashimoto, Kentaro Torisawa, Julien Kloetzer, Motoki Sano, Istv{\'a}n
  Varga, Jong-Hoon Oh, and Yutaka Kidawara.
\newblock 2014.
\newblock Toward future scenario generation: Extracting event causality
  exploiting semantic relation, context, and association features.
\newblock In {\em Proceedings of the 52nd Annual Meeting of the Association for
  Computational Linguistics (Volume 1: Long Papers)}, pages 987--997.

\bibitem[\protect\citename{Hu and Walker}2017]{hu2017inferring}
Zhichao Hu and Marilyn~A Walker.
\newblock 2017.
\newblock Inferring narrative causality between event pairs in films.
\newblock {\em arXiv preprint arXiv:1708.09496}.

\bibitem[\protect\citename{Kirkpatrick \bgroup et al.\egroup
  }1983]{Kirkpatrick1983OptimizationBS}
Scott Kirkpatrick, C.~D. Gelatt, and Mario~P. Vecchi.
\newblock 1983.
\newblock Optimization by simulated annealing.
\newblock {\em Science}, 220 4598:671--80.

\bibitem[\protect\citename{Luo \bgroup et al.\egroup
  }2016]{Luo2016CommonsenseCR}
Zhiyi Luo, Yuchen Sha, Kenny~Q. Zhu, Seung won Hwang, and Zhongyuan Wang.
\newblock 2016.
\newblock Commonsense causal reasoning between short texts.
\newblock In {\em KR}.

\bibitem[\protect\citename{Miller}1995]{miller1995wordnet}
George~A Miller.
\newblock 1995.
\newblock Wordnet: a lexical database for english.
\newblock {\em Communications of the ACM}, 38(11):39--41.

\bibitem[\protect\citename{Mintz \bgroup et al.\egroup
  }2009]{Mintz2009DistantSF}
Mike Mintz, Steven Bills, Rion Snow, and Dan Jurafsky.
\newblock 2009.
\newblock Distant supervision for relation extraction without labeled data.
\newblock In {\em ACL/IJCNLP}.

\bibitem[\protect\citename{Mirza and Tonelli}2014]{Mirza2014AnAO}
Paramita Mirza and Sara Tonelli.
\newblock 2014.
\newblock An analysis of causality between events and its relation to temporal
  information.
\newblock In {\em COLING}.

\bibitem[\protect\citename{Mirza and Tonelli}2016]{mirza2016catena}
Paramita Mirza and Sara Tonelli.
\newblock 2016.
\newblock Catena: Causal and temporal relation extraction from natural language
  texts.
\newblock In {\em The 26th international conference on computational
  linguistics}, pages 64--75. ACL.

\bibitem[\protect\citename{Oh \bgroup et al.\egroup }2013]{oh2013question}
Jong-Hoon Oh, Kentaro Torisawa, Chikara Hashimoto, Motoki Sano, Stijn
  De~Saeger, and Kiyonori Ohtake.
\newblock 2013.
\newblock Why-question answering using intra-and inter-sentential causal
  relations.
\newblock In {\em Proceedings of the 51st Annual Meeting of the Association for
  Computational Linguistics (Volume 1: Long Papers)}, pages 1733--1743.

\bibitem[\protect\citename{Oh \bgroup et al.\egroup }2017]{oh2017multi}
Jong-Hoon Oh, Kentaro Torisawa, Canasai Kruengkrai, Ryu Iida, and Julien
  Kloetzer.
\newblock 2017.
\newblock Multi-column convolutional neural networks with causality-attention
  for why-question answering.
\newblock In {\em Proceedings of the Tenth ACM International Conference on Web
  Search and Data Mining}, pages 415--424. ACM.

\bibitem[\protect\citename{Riaz and Girju}2014]{riaz2014recognizing}
Mehwish Riaz and Roxana Girju.
\newblock 2014.
\newblock Recognizing causality in verb-noun pairs via noun and verb semantics.
\newblock In {\em Proceedings of the EACL 2014 Workshop on Computational
  Approaches to Causality in Language (CAtoCL)}, pages 48--57.

\bibitem[\protect\citename{Schuler}2005]{schuler2005verbnet}
Karin~Kipper Schuler.
\newblock 2005.
\newblock Verbnet: A broad-coverage, comprehensive verb lexicon.

\bibitem[\protect\citename{Wei and Zou}2019]{wei-zou-2019-eda}
Jason Wei and Kai Zou.
\newblock 2019.
\newblock {EDA}: Easy data augmentation techniques for boosting performance on
  text classification tasks.
\newblock In {\em Proceedings of the 2019 Conference on Empirical Methods in
  Natural Language Processing and the 9th International Joint Conference on
  Natural Language Processing (EMNLP-IJCNLP)}, pages 6382--6388, Hong Kong,
  China, November. Association for Computational Linguistics.

\bibitem[\protect\citename{Wu \bgroup et al.\egroup }2016]{wu2016google}
Yonghui Wu, Mike Schuster, Zhifeng Chen, Quoc~V Le, Mohammad Norouzi, Wolfgang
  Macherey, Maxim Krikun, Yuan Cao, Qin Gao, Klaus Macherey, et~al.
\newblock 2016.
\newblock Google's neural machine translation system: Bridging the gap between
  human and machine translation.
\newblock {\em arXiv preprint arXiv:1609.08144}.

\bibitem[\protect\citename{Zuo \bgroup et al.\egroup }2020]{zuo2020causal}
Xinyu Zuo, Yubo Chen, Kang Liu, and Jun Zhao.
\newblock 2020.
\newblock Towards causal explanation detection with pyramid salient-aware
  network.

\end{thebibliography}

\end{document}